\documentclass[letterpaper, 10 pt, conference]{ieeeconf}

\IEEEoverridecommandlockouts

\usepackage{epsfig}
\usepackage{amsmath}
\usepackage{amssymb}
\usepackage[caption=false, font=footnotesize]{subfig}
\usepackage[dvipsnames]{xcolor}
\usepackage{multirow}
\usepackage[super]{nth}
\usepackage{hyperref}
\usepackage{makecell}
\usepackage{siunitx}
\usepackage[ruled,vlined]{algorithm2e}

\DeclareMathOperator{\sign}{sgn}

\begin{document}

\title{Enhancing Continuous Control of Mobile Robots for End-to-End Visual Active Tracking}

\author{Alessandro~Devo, Alberto~Dionigi, Gabriele~Costante
\thanks{The authors are with the Department of Engineering, University of Perugia, 06125 Perugia, Italy
        {\tt\footnotesize alessandro.devo@studenti.unipg.it, alberto.dionigi@studenti.unipg.it, gabriele.costante@unipg.it}}%
}

\maketitle

\begin{abstract}
In the last decades, visual target tracking has been one of the primary research interests of the Robotics research community. The recent advances in Deep Learning technologies have made the exploitation of visual tracking approaches effective and possible in a wide variety of applications, ranging from automotive to surveillance and human assistance. However, the majority of the existing works focus exclusively on \textit{passive} visual tracking, \textit{i.e.}, tracking elements in sequences of images by assuming that no actions can be taken to adapt the camera position to the motion of the tracked entity. On the contrary, in this work, we address visual \textit{active} tracking, in which the tracker has to actively search for and track a specified target. Current State-of-the-Art approaches use Deep Reinforcement Learning (DRL) techniques to address the problem in an end-to-end manner. However, two main problems arise: i) most of the contributions focus only on discrete action spaces and the ones that consider continuous control do not achieve the same level of performance; and ii) if not properly tuned, DRL models can be challenging to train, resulting in a considerably slow learning progress and poor final performance. To address these challenges, we propose a novel DRL-based visual active tracking system that provides continuous action policies. To accelerate training and improve the overall performance, we introduce additional objective functions and a Heuristic Trajectory Generator (HTG) to facilitate learning. Through an extensive experimentation, we show that our method can reach and surpass other State-of-the-Art approaches performances, and demonstrate that, even if trained exclusively in simulation, it can successfully perform visual active tracking even in real scenarios.
\end{abstract}

\IEEEpeerreviewmaketitle
\section{INTRODUCTION}

The capability to detect and track a target object across multiple frames collected by vision sensors, \textit{i.e.} Visual Tracking (VT), plays an important role in many Robotic researches. The possible applications of VT technologies span across different areas, \textit{e.g.}, autonomous driving, surveillance, robot manipulation and human assistance, to name a few.
In the last years, the advent of Deep Learning-based techniques has exponentially increased the progresses made in order to improve both performance and robustness of VT algorithms. In particular, the use of Convolutional Neural Networks (CNNs) has allowed the development of increasingly sophisticated and precise tracking systems. However, most of the works focus exclusively on tracking objects and/or people in pre-recorded videos \cite{nam2016learning, song2017crest, zhang2017deep, pu2018deep} or, in general, assume that the target is always within the field of view of a fixed camera, whose position cannot be adapted to the target motion. This condition considerably limits the possibility to apply VT approaches in many real-world robotic scenarios.

\begin{figure*}[t]
	\centering
	\includegraphics[width=1.8\columnwidth]{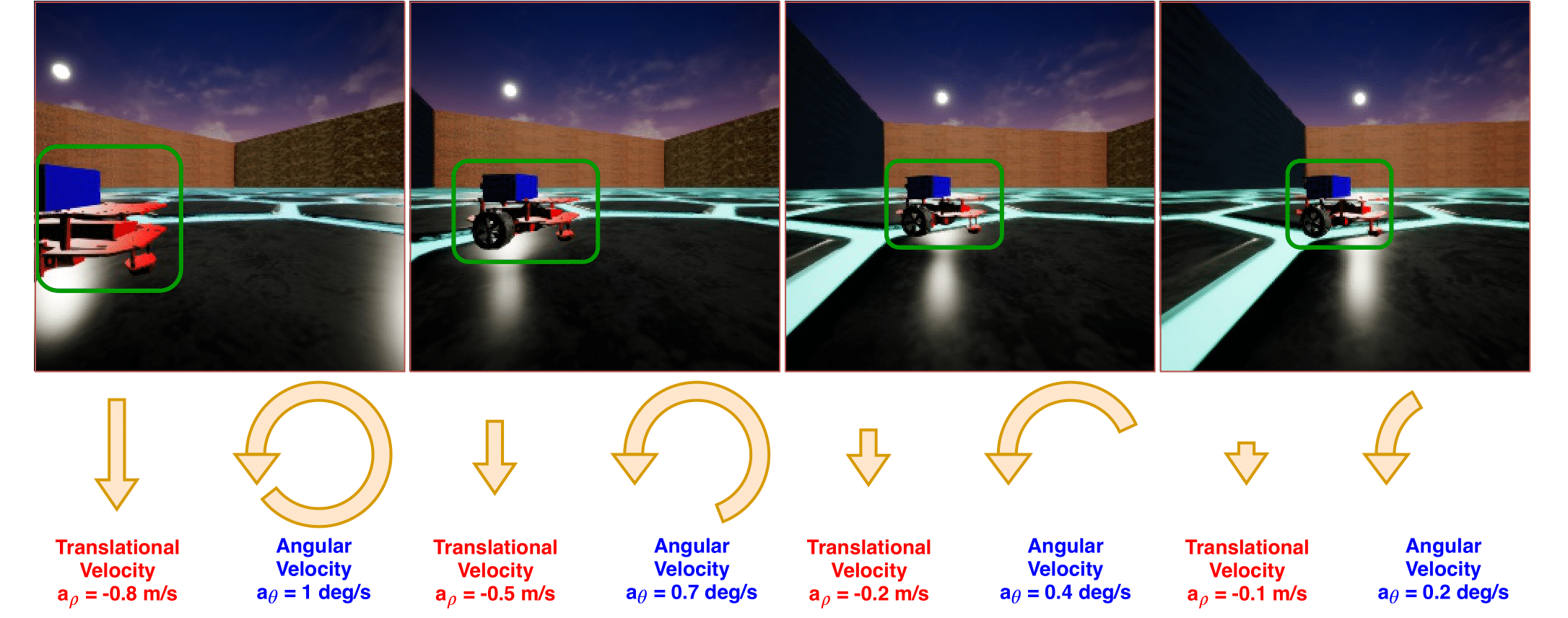}
	\caption{The visual active tracking task. The goal of the tracker is to maintain the target (marked in green) within its field of view. Contrary to passive tracking, which assume that the target is always within the field of view of a fixed camera, in the active scenario the tracker has to perform motion maneuvers to maintain view contact with it. In this example (from left to right), the tracker chooses to consecutively perform backward translation and left turning to center the target.}
	\label{task}
\end{figure*}

Therefore, in this work, we focus on the more challenging task of Visual Active Tracking (VAT) \cite{ccelik2017color, luo2019end, zhong2019ad}, in which the tracker has to \textit{actively} search for and track a specified target (Fig. \ref{task}). This task is clearly more complex than passive VT, since the tracker must not only identify the target but also act and change its position to maintain view contact with it. This requires the robot to localize itself with respect to the target and plan in real-time the most suitable trajectory to follow it.

A possible naive way to tackle the problem is to combine a generic object detection module, such as \cite{redmon2016you}, with other localization, collision detection and planning techniques \cite{mur2017orb, mancini2018j, engel2017direct, costante2020uncertainty, karaman2011sampling}. However, since this strategy is not specifically designed for target tracking, it is highly inefficient and generates a considerable information overhead.

A better solution is to get rid of the localization, mapping and collision detection blocks and focus only on two separate modules: the first has to identify and passively tracks the target in the images; the second is responsible for planning the tracker trajectory to follow the target \cite{murray1994motion, kim2005detecting, das2018stable}. Although this solution has brought interesting results, it still has some inefficiencies and the non-trivial problem of combining the two components remains.

For this reason, more recent works \cite{luo2019end, zhong2019ad} propose the use of Deep Reinforcement Learning (DRL) algorithms to address the problem in an end-to-end manner. DRL has proven to be remarkably effective in many vision, navigation and robotics tasks, particularly when multiple mutual objectives need to be accomplished simultaneously \cite{espeholt2018impala, andrychowicz2020learning, zhu2017target, devo2020deep, devo2020towards}. DRL-based VAT systems are able to process the image stream directly and provide action policies to adapt the camera field of view frame-by-frame, avoiding the need to manually adapt two separate vision and motion modules.

The well-known shortcomings of DRL algorithms, \textit{i.e.}, i) the need for a considerable amount of training samples for their optimization and ii) their trial-and-error learning-based nature have been addressed by relying upon synthetic environments \cite{luo2019end, zhong2019ad}. The availability of graphic engines with increased level of photorealism and the exploitation of recent sim-to-real domain adaptation techniques have made it possible to bridge the domain gap between DRL models trained on simulated scenarios and the real world without the need for fine-tuning procedures. In particular, among domain adaptation strategies, \textit{domain randomization} \cite{tobin2017domain} is certainly one of the most popular and has been successfully applied to various robotic tasks \cite{peng2018sim, andrychowicz2020learning} and even to VAT \cite{zhong2019ad}. However, most of these DRL approaches for VAT only consider discrete action spaces. On the other hand, continuous control methods are often difficult and slow to train, resulting in optimization procedures that require tens of millions of steps before starting to learn \cite{luo2019end} or may not even be successful.

For these reasons, in this work, we propose C-VAT, a novel DRL-based approach for VAT in \textit{continuous} action spaces, which features a new training procedure that produces sample efficient, effective and robust visual tracking policies. Specifically, we employ a deterministic tracking algorithm to considerably speed up the \textit{critic} component learning and an auxiliary loss that, since the beginning of training, helps the \textit{actor} to develop a useful basic understanding of the tracking task. Our approach is trained exclusively in simulated environments and benefits from domain randomization to achieve generalization over real world contexts. We evaluate our method performance in a large variety of synthetic experiments, and show that, even if trained with synthetic data only, our algorithm can be effectively used in real scenarios with physical robots.

This work proceeds as follows: Section \ref{related_work} contains our literature review; Section \ref{approach} formalizes the task, presents our approach and the environment setup; Section \ref{experiments} describes the experiments and shows the results; finally, Section \ref{conclusions} draws our conclusions and the path of future work.

\section{RELATED WORK}\label{related_work}

The majority of the state-of-the-art works focus on \textit{passive} visual tracking \cite{nam2016learning, song2017crest, zhang2017deep, pu2018deep} \textit{i.e.}, they assume that the camera cannot be moved to react to the target object motions and that the latter is always in the camera field of view. The authors in \cite{zhu2018distractor}, for example, propose a novel distractor-aware approach based on Siamese networks for visual object tracking. Conversely, in \cite{song2018vital}, an adversarial training technique is introduced. They perform data augmentation by using a generative neural network to generate useful masks. The one that maintains the best features is then identified by the network through adversarial learning. Finally, the authors of \cite{yun2017action} devise a novel DRL-based tracking approach that iteratively moves an initial bounding box to follow the target across the image sequence.

Despite the recent advances, the passive approach is still limited to cases in which the target is within the camera field of view. On the contrary, active tracking approaches cover a wider range of situations since they can take advantage of the mobility of the hardware in which they are installed, e.g., movable surveillance cameras or mobile robots. Various approaches \cite{murray1994motion, kim2005detecting, das2018stable} try to combine passive models with camera control modules to actively perform tracking. \cite{ribaric2004real} introduces a real-time visual tracking system for indoor human motion tracking, which separates the task in image acquisition and camera motion estimation, object motion detection and localization, and camera control. In \cite{bellotto2012cognitive}, a hierarchical system to control a set of surveillance cameras is presented. In \cite{ccelik2017color}, the authors propose a system composed by three main different components to perform mapping, detecting and tracking the object. \cite{hong2018virtual} introduces a new modular architecture that incorporates a model for perception and another one for the control policy. The authors used the former to produce a semantic image segmentation from the perceived RGB frame, which is then used by the latter to perform the actions.

While addressing the task with modular systems is possible, employing such solutions has many drawbacks. In particular, a flaw in one of the components can propagate to the entire system, causing an overall failure. Furthermore, combining the separates modules of visual tracking and camera control can be considerably expensive.

For these reasons, recent works are focusing on end-to-end solutions, in which a direct mapping between vision and motion is learned. To this end, DRL has shown to be particular effective in many visual-navigation \cite{jaderberg2016reinforcement, zhu2017target, devo2020deep, devo2020towards} and robotics tasks \cite{kober2013reinforcement, thuruthel2018model}. \cite{andrychowicz2020learning} proposes a DRL algorithm able to learn complex dexterous in-hand manipulation policies. \cite{mirowski2016learning} show how an agent can be trained, from raw image pixels, to effectively explore complex unknown mazes and find a specific target. \cite{levine2016end} proposes a system that, from raw image observations only, can directly control torques at the robots motors. In \cite{sadeghi2016cad2rl}, a system trained exclusively in simulation for collision-free indoor flight in real environments is designed.

Inspired by these successes, some works introduce DRL-based systems for end-to-end VAT \cite{li2020pose, luo2019end, zhong2019ad}. In particular, \cite{li2020pose} introduces a novel Pose-Assisted Multi-Camera Collaboration System composed by three main modules: the pose-based controller, the vision-based controller and a switcher that, in each step, chooses the best controller based on the visibility of the target. \cite{luo2019end} proposes a discrete action space CNN-LSTM model to track a human-like target, using only raw RGB frames as input. They also introduce a specifically designed reward function and an environment augmentation technique based on domain randomization \cite{tobin2017domain, peng2018sim, tremblay2018training, bousmalis2018using, andrychowicz2020learning} to generalize in real world scenarios. In addition, the authors propose a continuous variant of their method, which, however, due to the extra complexity of managing continuous actions, demonstrate lower performances that its discrete counterpart. Since in that work the target policy is fixed throughout all the learning process, \cite{zhong2019ad} introduces an asymmetric duelling training procedure, during which the target learn complex escape policies to avoid to be tracked. The authors demonstrate that such a mechanism allows the tracker to practice with a much more challenging target, making it more robust and faster to train.

\subsection{Contribution}
Current state-of-the-art VAT methods \cite{luo2019end, zhong2019ad} primarily focus on discrete action spaces, which limit their applicability do not allow them to achieve performances comparable to those of a continuous control approach. Therefore, we propose a novel training procedure based on the popular Deep Deterministic Policy Gradient (DDPG) algorithm \cite{lillicrap2015continuous}, which can successfully and consistently perform tracking with continuous actions. Due to the complexity of the task and the continuous action space and the \textit{off-policy} nature of the DDPG algorithm, other works that simply apply DRL to train continuous action policies obtain extremely poor results, as we demonstrate in our experimental section. On the contrary, our approach can effectively handle continuous action spaces, overcoming the limitations of previous state-of-the-art continuous control methods, and can produce robust tracking policies, surpassing also discrete action models performance.

To summarize, our contribution is three-fold:
\begin{enumerate}
	\item We introduce C-VAT, a DRL-based architecture that computes policies in continuous action space for end-to-end VAT;
	\item We devise a training technique that leverages different strategies to ease and speed-up the optimization process and achieve better performance than state-of-the-art baselines;
	\item We demonstrate that C-VAT, although trained exclusively on simulated environments, guarantees remarkable performance even in real scenarios, without the need for any tuning procedure.
\end{enumerate}

\section{APPROACH}\label{approach}

In this section, we first formally define the problem and provide a concise background on the classical RL setting. Secondly, we describe how we can formulate our specific VAT task as a DRL problem. In the third section, we explain in detail the proposed training procedure and network architecture. Finally, we presents the simulated environments we use to train our model.

\subsection{Problem Formulation}

\begin{figure}[t]
	\centering
	\includegraphics[width=0.7\columnwidth]{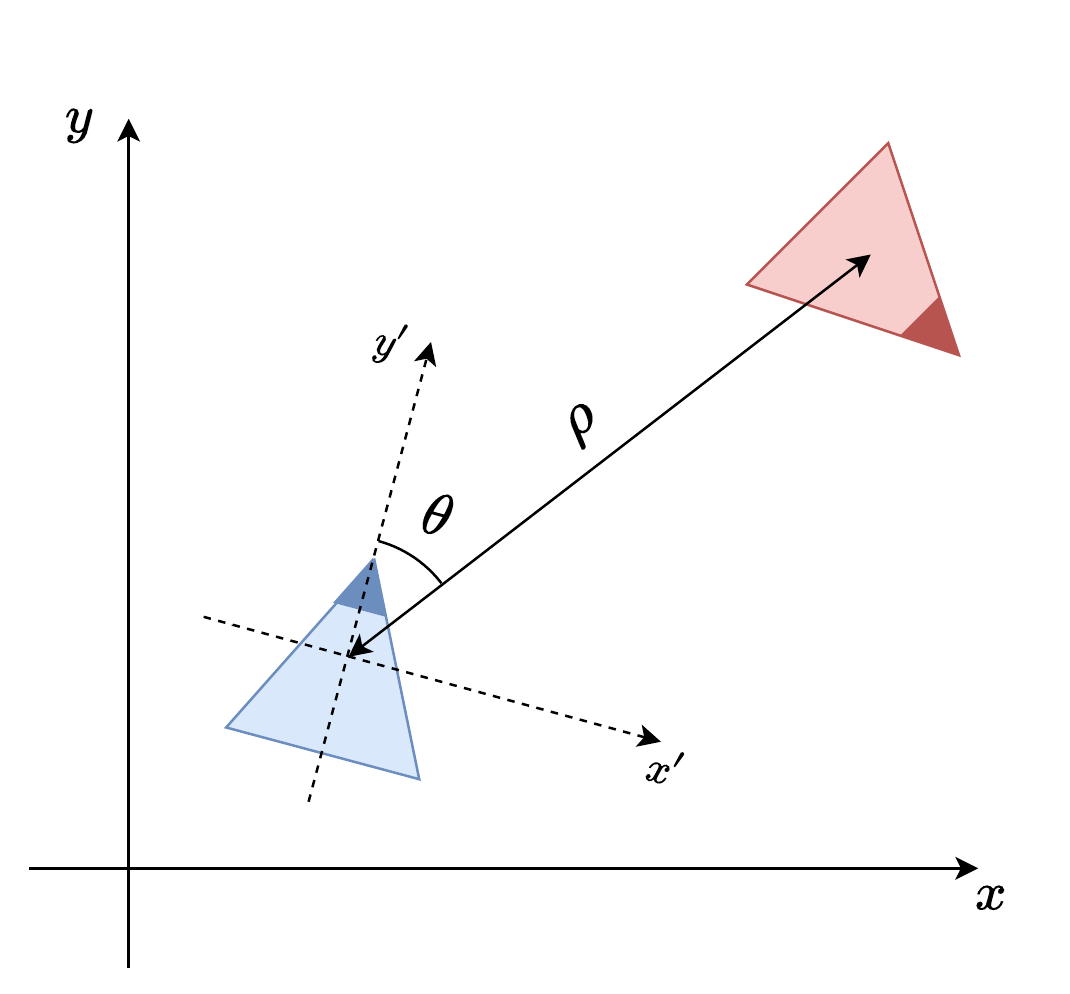}
	\caption{The considered coordinate system to compute the reward $r$. As can be observed, $\rho$ represents the euclidean distance between the \textit{tracker} (blue triangle) and the \textit{target} (red triangle), and $\theta$ constitutes the angle defined by $\rho$ and the \textit{tracker} forward direction. Both $\rho$ and $\theta$ are measured by considering the relative coordinate system w.r.t. the \textit{tracker}.}
	\label{coord_sys}
\end{figure}

The objective of an autonomous robot that performs VAT, referred to in the following as \textit{tracker}, is to recognize and actively track a predefined, and possibly moving, \textit{target}, by using only visual inputs. Formally, we frame the task as a classic RL problem \cite{sutton2018reinforcement}, in which an agent interacts with an environment $E$ over a discrete number of timesteps. The environment can be seen as a Markov Decision Process (MDP) in which the main task of the agent is to find a policy $\pi$ that maximizes the sum of discounted future reward:
\begin{equation}\label{R_eq}
	R_t=\sum_{i=t}^{T} \gamma^{i-t} r_i\left(x_i,a_i\right),
\end{equation}
where $\gamma \in [0,1)$ is the discount factor and $r_i(x_i,a_i)$ is the reward at time $i$, given the state $x_i$ and the action $a_i \sim \pi(\cdot|x_i)$.

We can also define the action-value function $Q_\pi$ as:
\begin{equation}\label{Q_eq}
	Q_{\pi}\left(x_t, a_t\right)=\mathbb{E}_{r_{i\geq t}, x_{i>t} \sim E, a_{i>t} \sim \pi} \left[ R_t | x_t, a_t \right],
\end{equation}
which describes the expected return after choosing an action $a_t$ in state $x_t$ and thereafter following policy $\pi$. It should be noticed that estimating correctly $Q_{\pi}$ means knowing exactly which is the best action $a_t$ to take for every $x_t$, and hence, solving the MDP.

It is important to remark, however, that the true state $x_t$ of the environment can be unknown to the agent (as in our setting). In such a case, $E$ is, in fact, a Partially Observable MDP (POMDP), which, instead, provides to the agent only an observation $o_t$ of the underlying state.

In the following, we define what $x_t$, $o_t$, $a_t$ and $r_t$ represent in our particular scenario (see Section \ref{env_and_task_details}) and describe the algorithm we employ to estimate $Q_\pi$ (see Section \ref{arch_train}).

\subsection{Task Details} \label{env_and_task_details}

In our specific case, we consider a VAT system whose inputs consist of the RGB frame collected by the \textit{tracker} camera (we assume it mounted in the front of the robot). Our \textit{tracker} is free to move along the X and the Y axes of a three-dimensional space by performing continuous actions, in order to adapt its position to keep the \textit{target} in its field of view. In this setting, we can notice that: i) a captured image represents only an observation $o_t$ of the unknown underlying state $x_t$ (\textit{i.e.,} the positions of the \textit{tracker} and the \textit{target}, and the map of the environment); ii) the observation space is extremely vast, since generated by all the possible combinations of pixel values; iii) the \textit{tracker} action space is infinite, since we consider continuous controls. For these reasons, classical RL algorithms cannot be applied and the use of more advanced DRL methods that exploit complex Deep Neural Network (DNN) approximators is necessary.

We start by considering a series of independent episodes during which our agent (\textit{i.e.,} the \textit{tracker}) interacts with the environment to collect visual observations, perform continuous control actions and get rewards. When an episode starts, the \textit{tracker} has to first look around to find the \textit{target}, since it can also spawn outside its initial field of view. Then, it can start to track it and maximize the reward signal $r_t$, which we defined as:
\begin{equation}\label{r_t_eq}
	r_t = Ar_{\rho_t}r_{\theta_t},
\end{equation}
where
\begin{equation}\label{e_rho_eq}
	r_{\rho_t} = \max\left(0, 1 - \frac{|\rho_t - \rho^*|}{\rho_{max}}\right),
\end{equation}
and
\begin{equation}\label{e_theta_eq}
	r_{\theta_t} = \max\left(0, 1 - \frac{|\theta_t - \theta^*|}{\theta_{max}}\right).
\end{equation}
In particular, in Eq. \eqref{e_rho_eq} $\rho_t$ represents the current distance between the \textit{tracker} and the \textit{target} while $\rho^*$ and $\rho_{max}$ the optimal and the maximum ones, respectively. Consequently, $r_{\rho_t}$ encodes how close the distance between the \textit{tracker} and the \textit{target} is to the optimal one. Conversely, in Eq. \eqref{e_theta_eq}, $r_{\theta_t}$ measures the \textit{tracker} ability to maintain the \textit{target} within its field of view, with $\theta_t$, $\theta^*$ and $\theta_{max}$ indicating the current, desired and maximum angular distances, respectively. All the angles are computed with respect to the center of the field of view, hence, $\theta^*$ is set to $\ang{0}$ if the \textit{tracker} is required to keep the \textit{target} in the center of the image (see Fig. \ref{coord_sys}).

From the equations, it can be noticed that $r_t$ is always included in the range $[0,A]$ (where $A$ is an algorithm hyperparameter) and is maximum when both $\rho_t$ and $\theta_t$ coincide exactly with $\rho^*$ and $\theta^*$, respectively. On the other end, $r_t$ is $0$ if $\rho_t$ exceeds $\rho_{max}$ or $\theta_t$ is greater than $\theta_{max}$.

The \textit{tracker} action space is continuous. At each timestep $t$, it produces two different and independent real valued actions, $a_{\rho_t}$ and $a_{\theta_t}$, which represents the translational and the angular speeds, respectively. It should be observed that such continuous actions allow extremely fine maneuvers that would not be possible with discrete ones. 

Regardless of the reward received and the actions performed, a training episode ends when a predefined number of steps is reached.

\begin{figure*}[t]
	\centering
	\includegraphics[width=1\linewidth]{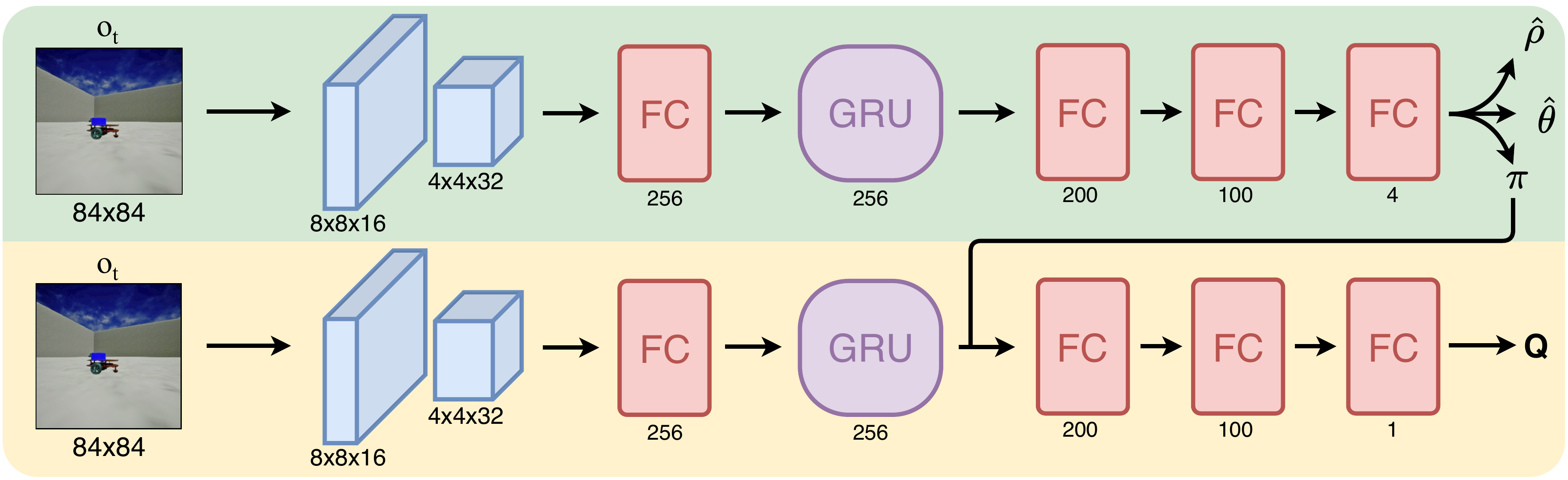}
	\caption{The proposed architecture for visual active tracking. It is composed by two main components: the \textit{actor} network (in green) and the \textit{critic} network (in yellow). Both are fed with the current $84\times84$ RGB frame, which is first processed by the actor to produce the estimated relative distance ($\hat{\rho}$) and angle ($\hat{\theta}$) for the auxiliary angle-distance loss, and the policy vector ($\pi$). The latter is then concatenated with the features extracted by the GRU of the critic, which, finally, outputs the Q-value function ($Q$). Despite the two networks are almost identical, their weights are not shared between them.}
	\label{gen_scheme}
\end{figure*}

During training, the \textit{target} position is kept fixed, hence, the \textit{tracker} job is simply to move itself in order to meet the desired $\theta^*$ and $\rho^*$ and, then, to maintain its current position and orientation. One could argue that, compared with other much more complex procedures (as those in used in \cite{zhong2019ad}), training with such a simple task could prevent the \textit{tracker} to generalize over more complex and general tracking scenarios, \textit{i.e.}, with moving targets. Nonetheless, in Section \ref{experiments}, we show that this simple training strategy is sufficient to learn robust tracking abilities for far more challenging scenarios. 

\subsection{Network Architecture and Training Algorithm} \label{arch_train}

To train our model we use the Asynchronous Advantage Actor Critic (A3C) \cite{mnih2016asynchronous} framework combined with the DDPG algorithm \cite{lillicrap2015continuous} (see Algorithm \ref{alg}), which is specifically designed to deal with continuous action spaces and DNN approximators.

We use several instances of our agent to collect trajectories of states, actions and rewards. All the copies have a local network, whose architecture is identical for all of them, which is periodically synchronised with a shared model. Each copy is placed in a different room, where it collects data and stores them in a personal \textit{replay buffer} \cite{wang2016sample}. Every time an agent has to be updated, a number of samples are randomly picked from its buffer. The agent uses these data to compute the losses (specific details on Sections \ref{ddpg_loss} and \ref{aux_loss}) and the gradients with respect to its local network. The gradients are then transferred to the shared network, which updates the parameters and sends them back to the local network. It should be noticed that since the agents are updated independently, the learning process is asynchronous and, due to the replay buffer, it is also off-policy.

\begin{algorithm*}[t]
\SetAlgoLined
Randomly initialize shared critic network with weights $W^Q$\\
Randomly initialize shared actor network with weights $W^\pi$\\
Initialize shared critic target network with weights $W^{Q'} \leftarrow W^Q$\\
Initialize shared actor target network with weights $W^{\pi'} \leftarrow W^\pi$\\
Initialize agent-specific critic network with weights $W^{Q''} \leftarrow W^Q$\\
Initialize agent-specific actor network with weights $W^{\pi''} \leftarrow W^\pi$\\
Initialize replay buffer\\
Initialize random process $\mathcal{N}$ for action exploration\\
 \For{episode = 1, M}{
  Receive initial observation $o_0$\\
  \For{t = 1, T}{
  	Select action $a_t=\pi\left(o_t|W^{\pi''}\right) + \mathcal{N}$ according to the current policy and exploration noise\\
  	Execute action $a_t$ and get reward $r_t$ and new observation $o_{t+1}$\\
  	Store transition $\left(o_t, a_t, r_t, o_{t+1}\right)$ in the replay buffer\\
  	\If{$t \mod U = 0$}{
  	Sample a random batch of $B$ transitions $\left(o_i, a_i, r_i, o_{i+1}\right)$ from the replay buffer\\
  	Set $y_i = r_i + \gamma Q'\left(o_{i+1}, \pi'_{i+1}\left(o_{i+1}|W^{\pi'}\right)|W^{Q'}\right)$\\
  	Compute agent-specific critic gradients $\partial W^{Q''}$ by minimizing: $\ell_Q = \frac{1}{B}\sum_i^B\left(y_i-Q\left(o_i, a_i|W^{Q''}\right)\right)^2$\\
  	Compute agent-specific actor gradients $\partial W^{\pi''}$ by minimizing: $\ell_{\pi} = \frac{1}{B}\sum_i^BQ\left(o_i, \pi_i\left(o_{i}|W^{\pi''}\right)|W^{Q''}\right)$\\
  	Copy agent-specific critic gradients in the shared critic network: $\partial W^Q \leftarrow \partial W^{Q''}$\\
  	Copy agent-specific actor gradients in the shared actor network: $\partial W^\pi \leftarrow \partial W^{\pi''}$\\
  	Perform asynchronous update of $W^Q$ and $W^\pi$ by using $\partial W^Q$ and $\partial W^\pi$ respectively\\
  	Update the shared critic target network: $W^{Q'} \leftarrow \tau W^Q + \left(1-\tau\right)W^{Q'}$\\
  	Update the shared actor target network: $W^{\pi'} \leftarrow \tau W^\pi + \left(1-\tau\right)W^{\pi'}$\\
  	Update the agent-specific critic network: $W^{Q''} \leftarrow W^Q$\\
  	Update the agent-specific actor network: $W^{\pi''} \leftarrow W^\pi$\\
  	}
  }
 }
 \caption{A3C-DDPG algorithm}
 \label{alg}
\end{algorithm*}

In the following sections, we explain in major detail the various elements that compose the proposed approach.

\subsubsection{Architecture Details}
Similarly to other \textit{actor-critic} algorithms, DDPG makes use of two main components: an \textit{actor}, which chooses the actions to be performed, and a \textit{critic}, which evaluates such actions. We implement these entities with two distinct DNNs, namely A-DNN for the \textit{actor} and C-DNN for the \textit{critic}, as shown in Fig. \ref{gen_scheme}. 

A-DNN is fed with a $84\times84$ RGB frame $\left(o_t\right)$ that is initially processed by $2$ convolutional layers: the first with $16$ $8\times8$ filters with stride $4$, and the second with $32$ $4\times4$ filters with stride $2$. Both layers are followed by a \textit{ReLU} activation and a \textit{GroupNorm} layer. The image features extracted are further elaborated by a fully connected layer followed by a Gated Recurrent Unit (GRU) \cite{cho2014properties} network, both with $256$ hidden nodes and \textit{ReLU} activations. We decide to implement a GRU over a Long Short-Term Memory (LSTM) \cite{hochreiter1997long} network since, in most applications, the difference in terms of performance is negligible \cite{chung2014empirical, cascianelli2018full}, however, the former has fewer parameters, translating in a reduced computational complexity. The GRU output is then fed to $2$ fully connected networks, with $200$ and $100$ neurons each and \textit{ReLU} activations. At this point, the A-DNN produces the policy vector $\pi=\left[\pi_{\rho},\pi_{\theta}\right]$ and the estimated distance $\hat{\rho}$ and angle $\hat{\theta}$ (which are discussed in the following sections), using a last dense layer with $4$ neurons and \textit{tanh} activation.

Since the vector $\pi$ is deterministic, to allow exploration of the MDP, a gaussian exploration noise $\mathcal{N}=\left[\mathcal{N}_{\rho}, \mathcal{N}_{\theta}\right]$, with mean $\mu$ and variance $\sigma^2$, is finally added to produce $a_{\rho}=\pi_{\rho}+\mathcal{N}_{\rho}$ and $a_{\theta}=\pi_{\theta}+\mathcal{N}_{\theta}$, which represent the speed and the angular speed, respectively, chosen by the actor. It is important to specify, that such exploration noise is applied during training only, and that, in test phase, the action vector $a=\left[a_{\rho},a_{\theta}\right]$ coincides with $\pi$.

The C-DNN structure mimics that one of the actor, except for the output of the GRU layer, which is concatenated with the vector $\pi$ (from the actor network), before feeding it to the fully connected layers. Finally, the scalar action-value $Q$ is produced by a last fully connected layer with linear activation.

\subsubsection{DDPG Losses}\label{ddpg_loss}
To update the network weights, we apply the standard losses of the DDPG algorithm. Hence, for the critic we have:
\begin{equation}\label{l_Q_eq}
	\ell_Q = \frac{1}{T}\sum_t^T\left(y_t-Q\left(o_t, a_t\right)\right)^2,
\end{equation}
where
\begin{equation}\label{y_t_eq}
	y_t = r_t + \gamma Q'\left(o_{t+1}, \pi'_{t+1} \right);
\end{equation}
and for the actor:
\begin{equation}\label{l_pi_eq}
	\ell_{\pi} = \frac{1}{T}\sum_t^TQ\left(o_t, \pi_t\right).
\end{equation}

Since performing the bootstrapping for the next observation $o_{t+1}$ with the same learned networks could lead to divergence, we create a copy of the actor and critic networks for calculating the target values $\pi'$ and $Q'$, respectively (Eq. \eqref{y_t_eq}). These targets are needed to stabilise learning, and their parameters ($w'$) are updated by slowly tracking the ones ($w$) of the learned networks: $w'\leftarrow \tau w + \left(1 - \tau\right)w'$, with $\tau \ll 1$. As already mentioned, the data are sampled from different replay buffers of size $n$, which serve the dual purpose of minimising correlations between samples and speed up learning. Further details on the DDPG algorithm can be found in the original paper \cite{lillicrap2015continuous}.

\subsubsection{Heuristic Trajectories}\label{HT}
Since, at the beginning of the episodes, the \textit{target} can be spawned outside of the field of view of the \textit{tracker}, as explained in Section \ref{env_and_task_details}, the reward observed by the agent can be considerably low and sparse. This is a common problem in RL and, especially for challenging applications like VAT, may entirely compromise the whole training process. To avoid that, we design an effective deterministic tracking algorithm that we refer to as Heuristic Trajectories Generator (HTG). Since it uses ground truth informations provided by the simulation engine, we utilize it to fill the replay buffers with very useful trajectories just from the beginning of training. In particular, it employs the following heuristic:
\begin{equation}\label{pi_b_theta_eq}
	\pi^h_{\theta} = -\min\left(\frac{2|\theta_t-\theta^*|}{\mbox{FOV}}, 1\right) \sign\left(\theta_t \right),
\end{equation}
and
\begin{multline}\label{pi_b_rho_eq}
\pi^h_{\rho} = \\
	\begin{cases}
		\min\left(\frac{|\rho_t - \rho^*|}{\rho_{max}}, 1\right) \sign\left(\rho_t - \rho^*\right), & \mbox{if } |\theta_t - \theta^*|<10 \\
		0, & \mbox{otherwise}
	\end{cases}
\end{multline}
where FOV is the field of view of the \textit{tracker}. During training, the actual action vector is then calculated by adding the gaussian exploration noise $\mathcal{N}$ to the vector $\pi^h=\left[\pi^h_{\rho},\pi^h_{\theta}\right]$: $a^h=\left[a^h_{\rho},a^h_{\theta}\right]=\left[\pi^h_{\rho}+\mathcal{N}_{\rho},\pi^h_{\theta}+\mathcal{N}_{\theta}\right]$. This is needed to guarantee that the heuristic trajectories contain enough variability to be valuable for training the critic. It should be noticed that, also because of the additive noise, the sequences of actions produced by this simple policy are not (and are not intended to be) optimal. However, as we show in Section \ref{experiments}, they demonstrate to be crucial for a successful training since remarkably useful for the critic network, which can start to evaluate reasonable policies right from the very beginning of training. It is also important to highlight that, contrary to Imitation Learning \cite{billard2008survey, argall2009survey}, in which example trajectories are used to directly push the learned policy toward a predefined, and possibly suboptimal, behaviour, this technique simply augments the trajectories distribution that the critic observes.

\begin{figure}[t]
	\centering
	\includegraphics[width=\columnwidth]{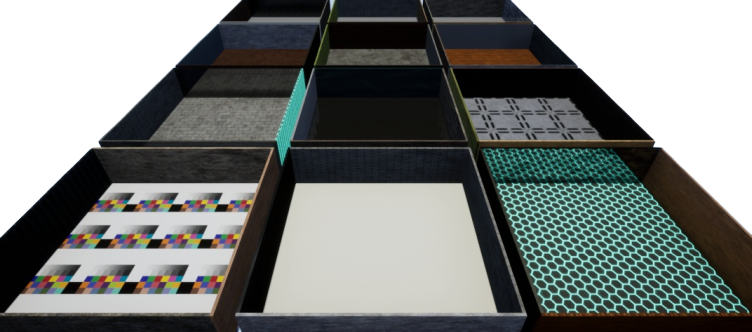}
	\caption{An overview of the training environments. Since we employ parallel training, each \textit{tracker-target} pair is placed inside one of the rooms. We employ a set of several different textures to randomize all the environments.}
	\label{overview}
\end{figure}

\subsubsection{Auxiliary Loss}\label{aux_loss}

\begin{figure*}[t]
	\centering
	\includegraphics[width=1.8\columnwidth]{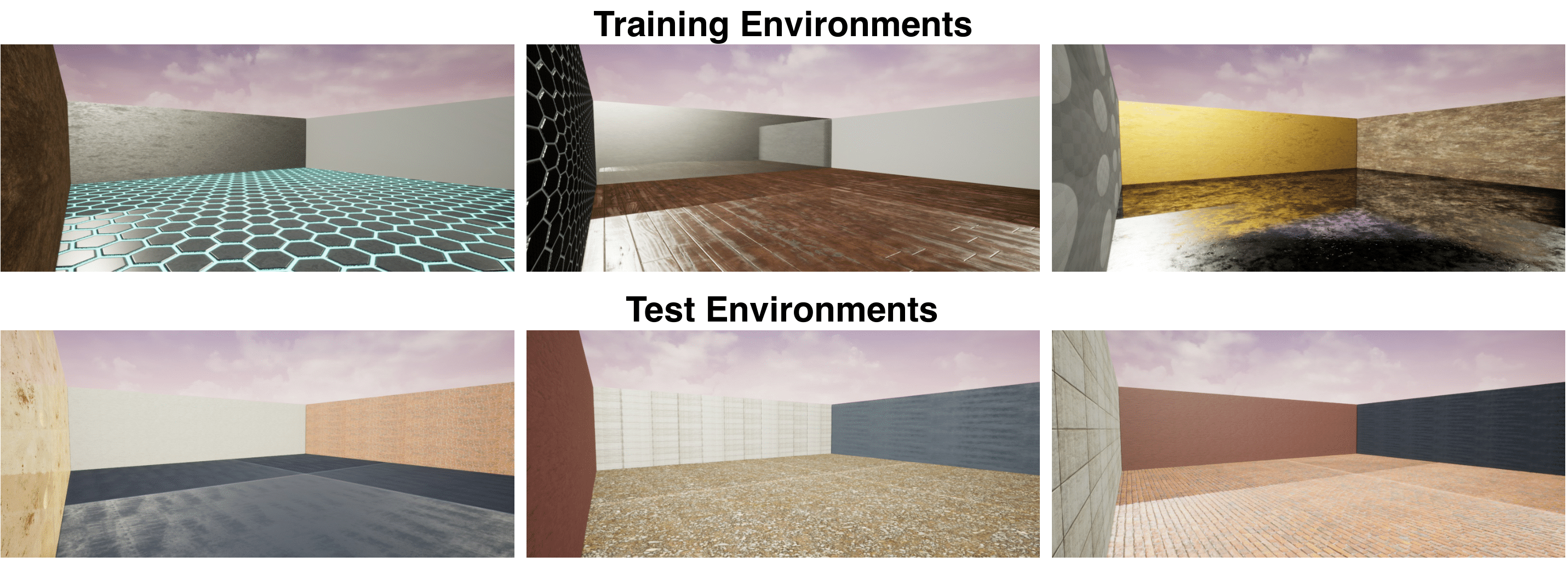}
	\caption{Some examples of randomized textures used in our training (top row) and test (bottom row) environments. The textures used during the learning phase differ significantly from those used for the evaluation.}
	\label{environments}
\end{figure*}

To help the agent learn basic tracking concepts, we add an auxiliary task in which the actor network has to estimate the relative angle and distance with respect to the \textit{target}. In particular, we introduce the following loss:
\begin{multline}\label{aux_eq}
	\ell_a = \frac{1}{L}\sum_t^L\left(\left(\hat{\rho}_t - \rho_t\right)^2 + \left(\hat{\theta}_t - \theta_t\right)^2 \right) + \\
	+ \frac{1}{L}\sum_t^L\left(\left(\hat{\rho}_t - \hat{\rho}_{t-1}\right)^2 + \left(\hat{\theta}_t - \hat{\theta}_{t-1}\right)^2 \right),
\end{multline}
where $L$ is the sequence length processed by the GRU, and $\hat{\rho}_t$ and $\hat{\theta}_t$ are, respectively, the distance and the angle estimated by the A-DNN at time $t$. The first term of the equation is the simple Mean Squared Error (MSE) between the estimated and the true values, while the second one is used to add temporal consistency to consecutive predictions.

To summarize, the overall optimization objective $\mathcal{L}$ for our C-VAT model can be written as:
\begin{equation}
	\mathcal{L} = \ell_Q + \ell_\pi + \ell_a.
\end{equation}

\subsection{Environment}

To train our model, we build a simulated environment (see Fig. \ref{overview}) using the photorealistic graphics engine Unreal Engine 4 (UE4)\footnote{https://www.unrealengine.com}. The environment consists of a large empty room, within which both the \textit{tracker} and the \textit{target} are positioned as follows: first, the \textit{tracker} is randomly spawned in the room and, afterward, the \textit{target} is randomly placed within a circumference of a predefined radius centered around the \textit{tracker}.

Since our agent is trained in simulated environments only, in order to achieve generalization also to real world contexts, we apply \textit{domain randomization} \cite{peng2018sim} to our synthetic scenario. This technique is successfully employed in many robotics and visual applications, such as: robotic harm control for object pushing \cite{peng2018sim}, object detection \cite{tremblay2018training} and robotic grasping \cite{bousmalis2018using}. It requires to randomize the training environment settings in order to make the system more robust to domain changes. Specifically, each time an episode ends we randomly change the lights conditions and the texture patterns of the rooms, including those of the walls, of the floor and of the \textit{target}.

\section{EXPERIMENTS} \label{experiments}

In the experiments, we aim to measure our C-VAT system performances, particularly w.r.t. environments not used for training. Specifically, we analyze our agent’s ability to: i) maintain the \textit{target} to the desired relative angle $\theta^*$ and distance $\rho^*$, and ii) to recover tracking in case the \textit{target} goes out-of-sight. To do that, we design several kind of tests, which are illustrated in the next sections. 

We compare and discuss the performance of C-VAT against two state-of-the-art baselines, the first one devised for discrete action spaces, while the second one for continuous control. We also propose an ablation study to evaluate the benefits brought by the HTG trajectories and the auxiliary angle-distance loss on the learning process and the final performance. Finally, to verify the generalization capability of our algorithm, we deploy the C-VAT model trained in the simulated environment (with no fine-tuning) in a robot within a real world scenario.

\begin{table}[t]
\renewcommand{\arraystretch}{1.3}
\caption{Settings and Hyperparameters}
\label{params}
\centering
\begin{tabular}{cc}
\hline
\textbf{Hyperparameter} & \textbf{Value} \\
\hline
Reward coeff. $\left(A\right)$ & $0.1$ \\
Optimal distance $\left(\rho^*\right)$ & $50$ cm \\
Max distance from optimal $\left(\rho_{max}\right)$ & $20$ cm \\
Optimal angle $\left(\theta^*\right)$ & $\ang{0}$ \\
Max angle from optimal $\left(\theta_{max}\right)$ & $\ang{10}$ \\
Exploration noise mean $\left(\mu\right)$ & $0$ \\
Exploration noise std $\left(\sigma\right)$ & $0.5$ \\
$\tau$ & $0.01$ \\
Number of agents & $10$ \\
Learning rate & $0.0001$ \\
Batch size $\left(B\right)$ & $128$ \\
Episode length & $250$ steps \\
Number of episodes $\left(M\right)$ & $9000$ \\
HTG episode length & $70$ steps \\
Replay buffer size per agent $\left(n\right)$ & $3,$$500$ trajectories \\
FOV & $\ang{90}$ \\
Agent speed & $\left[-8,8\right]$ cm/s \\
Agent steering speed & $\left[-8,8\right]$ deg/s \\
Sequence length $\left(L\right)$ & $5$ steps\\
Network update interval $\left(U\right)$ & $25$ steps \\
Discount factor $(\gamma)$ & $0.99$ \\
\hline
\end{tabular}
\end{table}

\begin{table*}[t]
\renewcommand{\arraystretch}{1.3}
\centering
\caption{Experimental Results in Simulated Environments}
\resizebox{2\columnwidth}{!}{
\label{syn_results}
\begin{tabular}{cc|ccccccccccccccccccc}
\cline{3-9}
\multirow{2}{*}{} & \multirow{2}{*}{} & \multirow{2}{*}{HTG+$\mathcal{N}$} & \multirow{2}{*}{HTG} & C-VAT & C-VAT & \multirow{2}{*}{AOT \cite{luo2019end}} & \multirow{2}{*}{AD-VAT+ \cite{zhong2019ad}} & \multirow{2}{*}{C-VAT}\\
&&&&(No HT-AL)&(No AL)&&&\\
\hline
\multirow{4}{*}{\textbf{\textit{dynamic\_target} $\boldsymbol{\left(3,3\right)}$}} & $\boldsymbol{p_{\rho}}$ & $0.90$ & $0.95$ & $0.14$ & $\boldsymbol{0.99}$ & $0.76$ & $0.83$ & $\boldsymbol{0.99}$\\
& $\boldsymbol{p_{\theta}}$ & $0.59$ & $0.64$ & $0.09$ & $\boldsymbol{0.94}$ & $0.73$ & $0.77$ & $\boldsymbol{0.94}$\\
& $\boldsymbol{p_c}$ & $0.75$ & $0.80$ & $0.12$ & $\boldsymbol{0.97}$ & $0.74$ & $0.80$ & $0.96$\\
& $\boldsymbol{p_v}$ & $0.98$ & $\boldsymbol{1.00}$ & $0.20$ & $\boldsymbol{1.00}$ & $0.79$ & $0.91$ & $\boldsymbol{1.00}$\\
\hline
\multirow{4}{*}{\textbf{\textit{dynamic\_target} $\boldsymbol{\left(5,5\right)}$}} & $\boldsymbol{p_{\rho}}$ & $0.72$ & $0.94$ & $0.15$ & $\boldsymbol{0.98}$ & $0.55$ & $0.86$ & $\boldsymbol{0.98}$\\
& $\boldsymbol{p_{\theta}}$ & $0.36$ & $0.41$ & $0.10$ & $\boldsymbol{0.89}$ & $0.51$ & $0.79$ & $\boldsymbol{0.89}$\\
& $\boldsymbol{p_c}$ & $0.54$ & $0.67$ & $0.12$ & $\boldsymbol{0.94}$ & $0.53$ & $0.83$ & $\boldsymbol{0.94}$\\
& $\boldsymbol{p_v}$ & $0.82$ & $\boldsymbol{1.00}$ & $0.22$ & $\boldsymbol{1.00}$ & $0.57$ & $0.93$ & $\boldsymbol{1.00}$\\
\hline
\multirow{4}{*}{\textbf{\textit{dynamic\_target} $\boldsymbol{\left(8,8\right)}$}} & $\boldsymbol{p_{\rho}}$ & $0.32$ & $0.29$ & $0.19$ & $\boldsymbol{0.94}$ & $0.12$ & $0.79$ & $\boldsymbol{0.94}$\\
& $\boldsymbol{p_{\theta}}$ & $0.19$ & $0.06$ & $0.14$ & $0.51$ & $0.10$ & $0.50$ & $\boldsymbol{0.65}$\\
& $\boldsymbol{p_c}$ & $0.25$ & $0.18$ & $0.17$ & $0.72$ & $0.11$ & $0.65$ & $\boldsymbol{0.80}$\\
& $\boldsymbol{p_v}$ & $0.39$ & $0.31$ & $0.30$ & $\boldsymbol{0.99}$ & $0.14$ & $0.85$ & $\boldsymbol{0.99}$\\
\hline
\multirow{4}{*}{\textbf{\textit{dynamic\_target} $\boldsymbol{\left(8,3\right)}$}} & $\boldsymbol{p_{\rho}}$ & $0.31$ & $0.31$ & $0.08$ & $\boldsymbol{0.87}$ & $0.17$ & $0.67$ & $\boldsymbol{0.87}$\\
& $\boldsymbol{p_{\theta}}$ & $0.21$ & $0.21$ & $0.05$ & $\boldsymbol{0.87}$ & $0.15$ & $0.65$ & $\boldsymbol{0.87}$\\
& $\boldsymbol{p_c}$ & $0.26$ & $0.26$ & $0.06$ & $\boldsymbol{0.87}$ & $0.16$ & $0.66$ & $\boldsymbol{0.87}$\\
& $\boldsymbol{p_v}$ & $0.85$ & $0.66$ & $0.15$ & $\boldsymbol{0.95}$ & $0.20$ & $0.77$ & $0.94$\\
\hline
\multirow{4}{*}{\textbf{\textit{random\_target}}} & $\boldsymbol{p_{\rho}}$ & $0.84$ & $0.93$ & $0.05$ & $0.91$ & $0.85$ & $0.74$ & $\boldsymbol{0.98}$\\
& $\boldsymbol{p_{\theta}}$ & $0.77$ & $0.84$ & $0.03$ & $0.90$ & $0.82$ & $0.79$ & $\boldsymbol{0.96}$\\
& $\boldsymbol{p_c}$ & $0.80$ & $0.88$ & $0.04$ & $0.91$ & $0.83$ & $0.77$ & $\boldsymbol{0.97}$\\
& $\boldsymbol{p_v}$ & $0.99$ & $\boldsymbol{1.00}$ & $0.10$ & $0.94$ & $0.88$ & $0.86$ & $\boldsymbol{1.00}$\\
\hline
\multirow{4}{*}{\textbf{\textit{adversarial\_target}}} & $\boldsymbol{p_{\rho}}$ & $0.52$ & $0.55$ & $0.06$ & $0.78$ & $0.08$ & $0.51$ & $\boldsymbol{0.80}$\\
& $\boldsymbol{p_{\theta}}$ & $0.45$ & $0.49$ & $0.05$ & $0.78$ & $0.07$ & $0.35$ & $\boldsymbol{0.80}$\\
& $\boldsymbol{p_c}$ & $0.49$ & $0.52$ & $0.06$ & $0.78$ & $0.07$ & $0.43$ & $\boldsymbol{0.80}$\\
& $\boldsymbol{p_v}$ & $0.73$ & $0.71$ & $0.16$ & $0.88$ & $0.10$ & $0.58$ & $\boldsymbol{0.90}$\\
\hline
\end{tabular}}
\end{table*}

\subsection{Implementation Details}

\begin{figure}[t]
	\centering
	\includegraphics[width=\columnwidth]{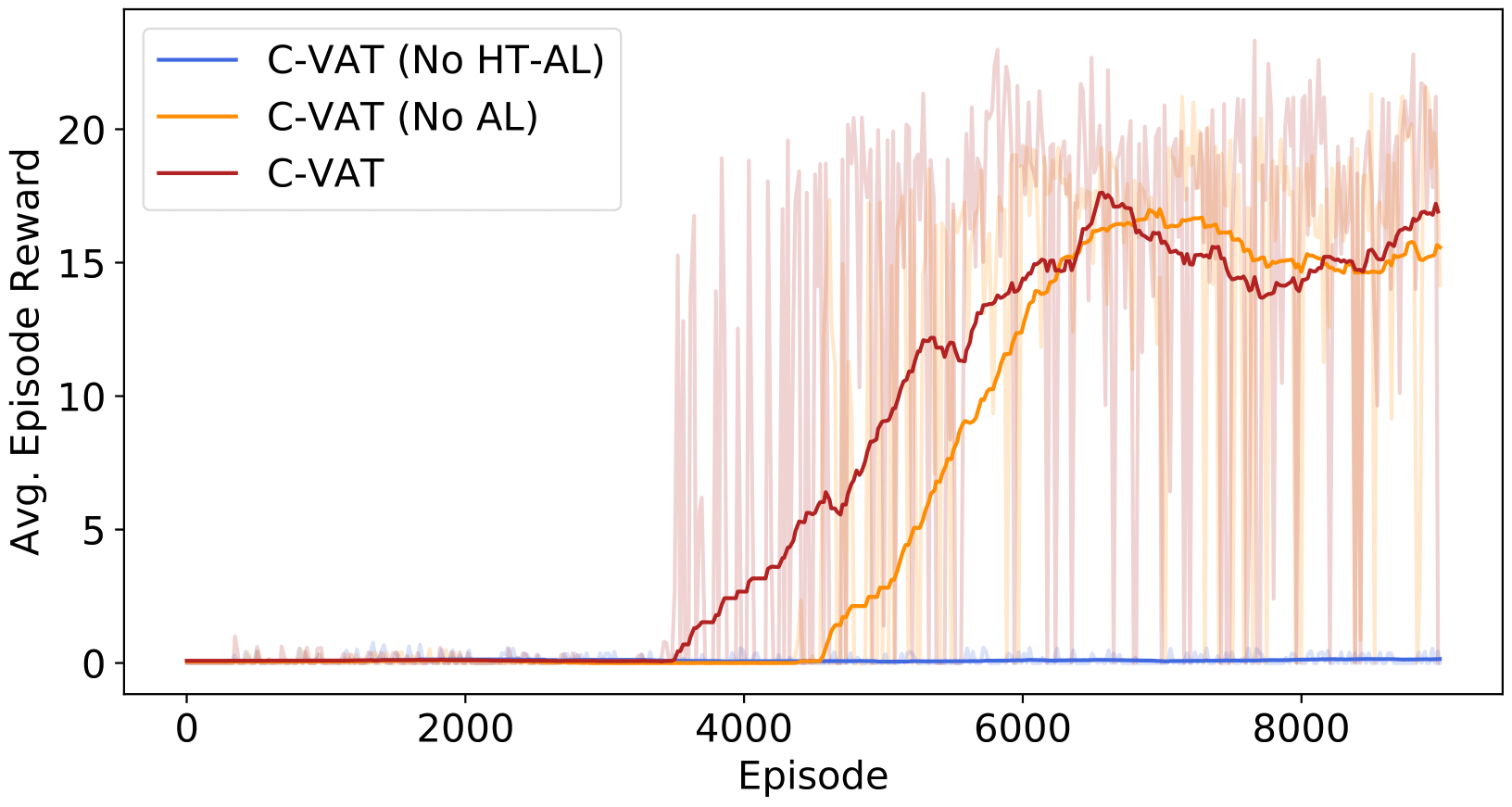}
	\caption{Average episode reward during training. The figure shows that the heuristic trajectories are necessary to the development of an effective policy. \textbf{C-VAT (No AL)} and \textbf{C-VAT} approximately reach the same level of performance, however, the plot demonstrates that the auxiliary angle-distance loss is helpful to speed up the learning process.}
	\label{episode_reward}
\end{figure}

We train both networks with $10$ A3C \cite{mnih2016asynchronous} agents for $9,$$000$ episodes using the Adam optimizer, with initial learning rate of $0.0001$ and by setting the batch size to $128$. The episodes performed by the A-DNN are composed by $250$ steps, while the episodes collected using the HTG have length $70$. In order to compensate for the episode length difference, we insert the trajectories in the replay buffer with a ratio of $\frac{1}{4}$ between the trajectories of the actor and those generated by the HTG.

All the experiments are run by using a workstation equipped: $2$ $\times$ NVIDIA GTX $2080$Ti with $11$GB of VRAM, Intel Core processor i$7$-$9800$X ($3.80$GHz $\times$ $16$) and $64$GB of DDR$4$ RAM. With this configuration, our model training takes about $140$ minutes (roughly $1s$ per episode), while, during the test phase, its forward pass requires $2ms$ per frame on average.

The specific training and environment parameter settings that are used throughout our experiments are given in detail in Table \ref{params}. For a better understanding of the testing scenarios, several examples of simulated and real experiments can be examined in the video attachment.

\subsection{Evaluation Scenarios, Metrics and Models}

To demonstrate the validity of our approach, we run a large variety of simulated experiments in a new set of previously unknown environments (Fig. \ref{environments}). We also compare our C-VAT method against $6$ baselines:
\begin{itemize}
	\item \textbf{HTG+$\boldsymbol{\mathcal{N}}$}: the algorithm used during training to collect the heuristic trajectories;
	\item \textbf{HTG}: the algorithm used during training to collect the heuristic trajectories, without the exploration noise $\mathcal{N}$;
	\item \textbf{C-VAT (No HT-AL)}: the basic version of our approach (A3C + DDPG);
	\item \textbf{C-VAT (No AL)}: the model that is trained by using also the heuristic trajectories from the HTG but without the auxiliary distance-angle loss $\ell_a$ (Eq. \eqref{aux_eq});
	\item \textbf{Active Object Tracking (AOT)}: the state-of-the-art approach proposed in \cite{luo2019end}. We consider the continuous action space variant, since it is more similar to our approach than its discrete counterpart. As in the original paper, the model is trained against a target that follows randomly generated trajectories, which we refer to as \textit{random\_target};
	\item \textbf{AD-VAT+}: the state-of-the-art approach proposed in \cite{zhong2019ad}. As in the original work, the model policy is learned during the adversarial duelling training.
\end{itemize}
\textbf{C-VAT (No HT-AL)}, \textbf{C-VAT (No AL)} and \textbf{C-VAT} are all tested without the exploration noise $\mathcal{N}$, and, as \textbf{AOT}, can move at a speed and steering speed within the range $\left[-8,8\right]$ cm/s and $\left[-8,8\right]$ deg/s, respectively. Conversely, \textbf{AD-VAT+} can only perform the discrete actions, with speed $8$ cm/s and steering speed $8$ deg/s, in the set: \{\textit{move-forward}, \textit{move-backward}, \textit{turn-left}, \textit{turn-right}, \textit{turn-left-and-move-forward}, \textit{turn-right-and-move-forward}, \textit{no-op}\}, as explained in \cite{zhong2019ad}.

To evaluate the agents performances, we design $3$ different kinds of simulated experiments:
\begin{itemize}
	\item \textit{dynamic\_target}: the \textit{target} is spawned in front of the \textit{tracker} and performs a circular trajectory in alternating random directions;
	\item \textit{random\_target}: the \textit{target} is spawned in front of the \textit{tracker} and performs randomly generated trajectories;
	\item \textit{adversarial\_target}: the \textit{target} is spawned in front of the \textit{tracker} and follows the policy learned during the adversarial duelling training \cite{zhong2019ad};
\end{itemize}

and $4$ metrics:
\begin{equation}
	p_{\rho_t} = 
	\begin{cases}
		\max\left(0, 1 - \frac{|\rho_t - \rho^*|}{150}\right), & \mbox{if } |\theta_t - \theta^*| > \frac{\mbox{FOV}}{2} \\
		0, & \mbox{otherwise}
	\end{cases},
\end{equation}
\begin{equation}
	p_{\theta_t} = 
	\begin{cases}
		\max\left(0, 1 - \frac{2|\theta_t - \theta^*|}{\mbox{FOV}}\right), & \mbox{if } |\rho_t - \rho^*| > 150 \\
		0, & \mbox{otherwise}
	\end{cases},
\end{equation}
\begin{equation}
	p_{c_t} = \frac{p_{\rho_t} + p_{\theta_t}}{2},
\end{equation}
\begin{equation}
	p_{v_t} =
	\begin{cases}
		1, & \mbox{if } p_{c_t}>0 \\
		0, & \mbox{otherwise}
	\end{cases}.
\end{equation}

The first $\left(p_{\rho_t}\right)$ and second $\left(p_{\theta_t}\right)$ metrics measure the \textit{tracker} ability to maintain the specified distance and angle, respectively, from the \textit{target}. It should be noted that both scores are $0$ if the \textit{target} exceeds the allowed distance $\left(150 \mbox{ cm}\right)$ or angle $\left(\frac{\mbox{FOV}}{2}\right)$. $p_{c_t}$ is simply the average of the first two metrics, and $p_{v_t}$ represents the percentage of steps in which the \textit{tracker} has the \textit{target} in view. All these metrics are averaged over the number of steps per run and by the number of runs.

\begin{figure*}[t]
	\centering
	\includegraphics[width=2\columnwidth]{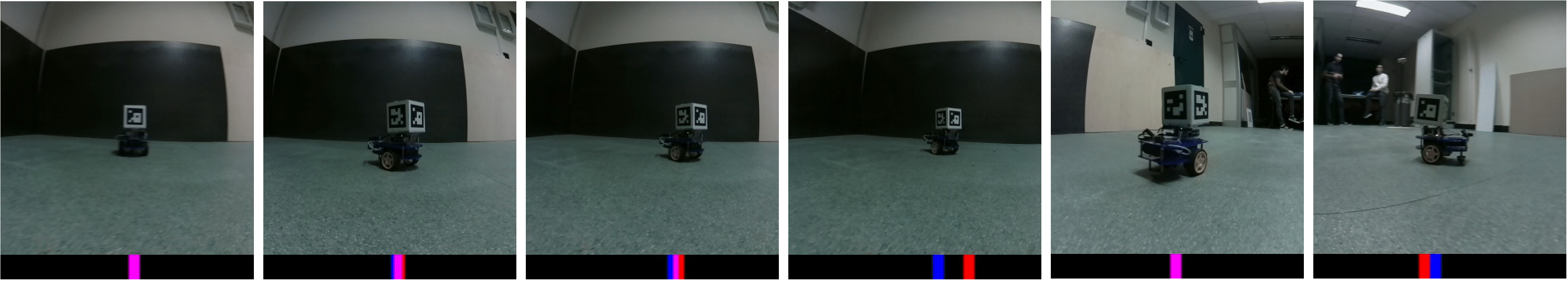}
	\caption{Six images captured by the camera mounted on the \textit{tracker} robot during the real world experiments. At the bottom of each figure, the blue and the red bars represent the current angle estimated by \textbf{C-VAT} and the one computed by the Aruco detector \cite{garrido2014automatic}, respectively. As it can be observed, the two estimates are very similar and often coincide. The fact that, in some cases, the difference between them increases can be explained by noticing that the Aruco markers are not placed at the center of the \textit{target}. It is important to highlight that \textbf{C-VAT} does not use any sort of marker to perform tracking (neither for training nor for testing). They are only used by the Aruco detector in order to provide a quantitative evaluation of our approach performance.}
	\label{real_angle}
\end{figure*}

\begin{figure}[t]
	\centering
	\includegraphics[width=\columnwidth]{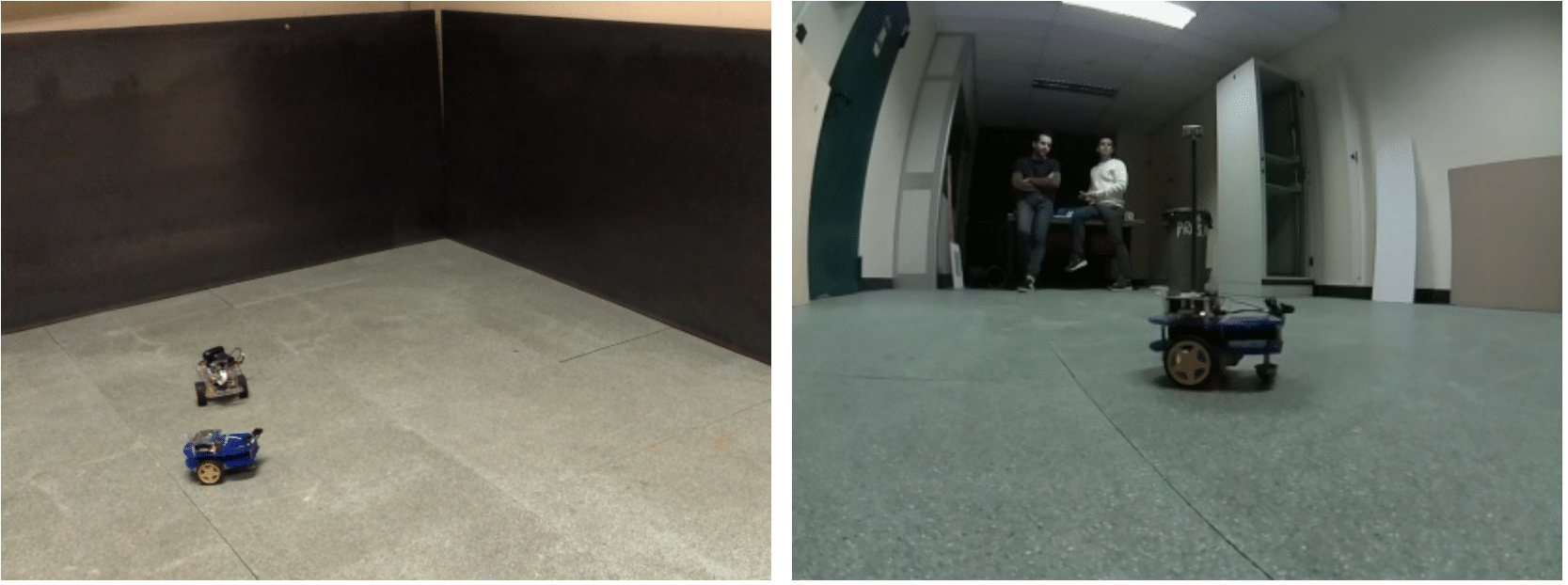}
	\caption{Two images captured from the recording camera (left) and the \textit{tracker} robot point of view (right) during the real world experiments. The visual appearance of our indoor environment is extremely different from that one of the simulated scenarios used during training.}
	\label{real}
\end{figure}

All the experiments have the same fixed length duration of $250$ steps. In order to evaluate the models performance for different \textit{target} velocities, in the \textit{dynamic\_target} scenario, we choose to vary its speed and steering speed within the set $\{\left(3, 3\right),\left(5, 5\right),\left(8, 8\right),\left(8, 3\right)\}$. Since the \textit{adversarial\_target} experiments employs a \textit{target} trained with a fixed speed and steering speed (equals to $8$ cm/s and $8$ deg/s, respectively), we decide to maintain the same values also for testing. All the aforementioned scenarios are averaged over $20$ runs.

\subsection{Results - Synthetic Experiments}

The results of the simulated experiments are summarized in Table \ref{syn_results}. As can be observed, our approach is the one that shows the best overall performances. In particular, it behaves very similarly to \textbf{C-VAT (No AL)}, which, however, in some scenarios (see \textit{dynamic\_target} $\left(8,8\right)$, \textit{random\_target} and \textit{adversarial\_target}), appears to be less accurate, as reflected by both $p_{\rho}$ and $p_{\theta}$ metrics. From the poor results of \textbf{C-VAT (No HT-AL)}, it is evident that training with heuristic trajectories has contributed significantly to the development of our tracker capabilities. This is also confirmed by the curves in Fig. \ref{episode_reward}, which represent our models' average episode reward throughout training.

Both \textbf{C-VAT (No AL)} and \textbf{C-VAT} outperform \textbf{AD-VAT+} in all scenarios and metrics. This may be due to the continuous action space they features, which allows them to track the \textit{target} more accurately. Interestingly, the larger gap is observed with the \textit{adversarial\_target}, which proves capable to systematically elude \textbf{AD-VAT+}. By qualitatively analyzing the experiments (see the video attachment), the \textit{adversarial\_target} seems to have developed a clever policy that exploits the fact that \textbf{AD-VAT+} cannot simultaneously move backward and turning. This strategy, however, is much less effective against \textbf{C-VAT}, which benefits from a symmetrical action space.

Our approach shows much better performance also with respect to \textbf{AOT}, especially against the targets with higher speeds (\textit{dynamic\_target} $\left(8,8\right)$, \textit{dynamic\_target} $\left(8,3\right)$ and \textit{adversarial\_target}). This is caused by the fact that during training \textbf{AOT} does not learn to lower its entropy in action selection and consequently the tracking policy appears quite ``noisy''. Because of that, it becomes rather common for the tracker to lose its target when it moves very quickly.

Remarkably, \textbf{C-VAT} dramatically outperforms also the \textbf{HTG}, both the noisy and non-noisy variants. As already explained in Section \ref{HT}, the heuristic trajectories are only needed in the early training stage, where the reward is still sparse. Indeed, as the results demonstrate, the \textbf{C-VAT} learned policy is not limited by the \textbf{HTG} trajectories.

\subsection{Results - Real Experiments}

\begin{table}[t]
\renewcommand{\arraystretch}{1.3}
\centering
\caption{Experimental Results in Real Environments}
\label{real_results}
\begin{tabular}{c|cccc}
\hline
\textbf{Run} & $\boldsymbol{p_{\rho}}$ & $\boldsymbol{p_{\theta}}$ & $\boldsymbol{p_c}$ & $\boldsymbol{p_v}$ \\
\hline
$1$ & $0.72$ & $0.85$ & $0.78$ & $0.89$ \\
$2$ & $0.74$ & $0.83$ & $0.79$ & $0.89$ \\
\hline
\end{tabular}
\end{table}

With these experiments, we want to asses \textbf{C-VAT} generalization capabilities and its robustness in real conditions. To this aim, we deploy our algorithm, without any kind of fine-tuning, in an indoor environment on a real two-wheeled mobile robot (Fig. \ref{real_angle}, \ref{real}). Both the \textit{tracker} and the \textit{target} robots can perform the same actions as their simulated counterparts. The former is controlled by our \textbf{C-VAT} model residing in a remote host, while the latter is manually controlled. Since, in the real world, we do not have access to ground truth informations, we equip the \textit{target} robot with Aruco markers and use the Aruco detector algorithm \cite{garrido2014automatic} to approximate the real relative angle and distance from the \textit{tracker}. In frames where measurements could not be obtained, we decide to calculate them by linear interpolation with those of adjacent frames. It is important to specify that the markers are used only for evaluation, since our method does not need them to perform tracking.

All the numerical results are reported in Table \ref{real_results}. Although \textbf{C-VAT} exhibits lower performance than those in simulation, it is still able to achieve remarkable results also in a real world environment. It should be noticed that this test scenario significantly differs from the simulated ones in terms of visual appearance. In particular, it is characterized by various objects, people, textures and lightning conditions that are completely absent in the training environments. Despite that, in Fig. \ref{real_angle}, which shows the model angle estimation performances, it is shown that the robot is actually able to recognize and locate the target.

Additional qualitative results of these experiments are available in the attached video.

\section{CONCLUSIONS} \label{conclusions}

In this work, we presented a new DRL-based approach for Visual Active Tracking that deals with continuous action spaces. Through extensive experimentation, we showed that our approach can perform robust tracking in both synthetic and real environments. Despite the relevant complexity of training a continuous control policy for VAT, our novel learning procedure demonstrated to be able to produce an end-to-end tracking model more accurate and reliable than other state-of-the-art systems.

Although the results are promising, there are still many open problems and aspects to improve. Specifically, we intend to continue our work by addressing more complicated scenarios, such as multi-target tracking, which introduces new challenges and requires more sophisticated methods.

\end{document}